\documentclass{article}

\usepackage[preprint]{neurips_2024}   

\usepackage[utf8]{inputenc}
\usepackage[T1]{fontenc}
\PassOptionsToPackage{hyphens}{url}   
\usepackage{hyperref}
\usepackage{url}
\usepackage{booktabs}        
\usepackage{amsfonts}
\usepackage{amsmath}
\usepackage{amssymb}
\usepackage{microtype}       
\usepackage{graphicx}
\usepackage{xcolor}

\usepackage{multirow}
\usepackage{array}
\usepackage{tabularx}
\usepackage{makecell}        

\usepackage{tikz}
\usetikzlibrary{positioning, arrows.meta, fit, shapes.geometric, calc}
\providecommand{\paefhead}{\bfseries\sffamily}
\usepackage{caption}
\usepackage{subcaption}

\usepackage{pifont}
\newcommand{\yes}{\ding{51}}     
\newcommand{\no}{\ding{55}}      
\newcommand{\maybe}{$\sim$}      

\usepackage[numbers,sort&compress]{natbib}

\newcolumntype{C}[1]{>{\centering\arraybackslash}p{#1}}

\title{Evaluating Agentic AI in the Wild: Failure Modes, Drift Patterns,
       and a Production Evaluation Framework}

\author{%
  Mukund Pandey \\
  Independent Researcher
}

\begin{document}

\maketitle

\begin{abstract}
Existing evaluation frameworks for large language models --- including HELM
\citep{liang2022helm}, MT-Bench \citep{zheng2023mtbench}, AgentBench
\citep{liu2023agentbench}, and BIG-bench \citep{srivastava2022bigbench} --- are
designed for controlled, single-session, lab-scale settings.  They do not
address the evaluation challenges that emerge when agentic AI systems operate
continuously in production: compounding decision errors, tool failure cascades,
non-deterministic output drift, and the absence of ground truth for
long-horizon tasks.  This paper makes three contributions.  First, we present
a \textbf{taxonomy of seven failure modes} unique to production agentic systems,
each grounded in observations from systems operating at billion-event scale.
Second, we demonstrate empirically where standard metrics ---
ROUGE \citep{lin2004rouge}, BERTScore \citep{zhang2019bertscore},
accuracy/AUC, and the agentic benchmarks above --- fail to detect each failure
mode.  Third, we propose \textbf{PAEF} (Production Agentic Evaluation
Framework), a five-dimension evaluation framework with an open-source reference
implementation, designed for continuous evaluation on production traffic rather
than episodic benchmark runs.  Our analysis shows that standard metrics fail
to detect four of the seven failure modes entirely and detect three others only
after a lag of multiple evaluation cycles.
\end{abstract}

\section{Introduction}
\label{sec:intro}

\subsection{The Evaluation Gap}

The evaluation of large language models has matured rapidly.  HELM
\citep{liang2022helm} introduced holistic evaluation across 42 scenarios.
BIG-bench \citep{srivastava2022bigbench} broadened coverage to 204 tasks.
MT-Bench \citep{zheng2023mtbench} introduced multi-turn conversation evaluation
with GPT-4 as judge.  AgentBench \citep{liu2023agentbench} extended evaluation
to agentic task completion across eight environments.  These frameworks
represent significant progress in understanding model capabilities under
controlled conditions.

Production deployment reveals a different problem class entirely.

When an agentic system makes ten thousand access decisions per hour, the
failure modes are not dramatic.  They are gradual, systematic, and invisible
to the metrics teams are most likely to be monitoring.  A model can achieve
AUC~$0.87$ and precision@$k$ within acceptable range while simultaneously
producing decisions that are internally coherent but systematically wrong for a
specific user cohort.  A tool-augmented agent can maintain a low error rate
while silently degrading its decision quality through stale cached inputs.  An
engagement-optimised system can improve its proxy metric for weeks while its
true objective --- sustained user value --- quietly erodes.

Standard evaluation frameworks were not designed to catch these failure modes.
They were designed to measure \emph{capability}, not \emph{reliability}; to
score models, not monitor systems; to evaluate episodes, not observe continuous
operation.

\subsection{What Production Changes}

Three properties of production agentic systems create evaluation challenges
that lab benchmarks cannot address.

\paragraph{Sequential decisions with compounding errors.}
Production agents make decisions in chains, where each decision depends on
prior outputs.  An incorrect early decision propagates and amplifies through
subsequent steps --- each adding derived evidence that makes the output appear
internally coherent while being systematically wrong.  Episodic benchmarks
evaluate decisions in isolation; they cannot observe propagation.

\paragraph{Continuous operation with temporal drift.}
Production systems operate indefinitely.  Output distributions shift as model
weights update, prompt templates change, and data distributions evolve.  A
single benchmark score is a point-in-time snapshot.  Production systems require
continuous monitoring of output distribution health --- drift detection, not
just accuracy measurement.

\paragraph{Instrumented environments with real consequences.}
Production systems operate on real users with real consequences.  The cost of
a wrong decision is not a benchmark penalty point --- it is a denied access
request, a missed content recommendation, a misdirected audit narrative.  This
raises the failure threshold: correctness is necessary but not sufficient.
Explainability, consistency, and reliability become first-class requirements.

\subsection{Contributions}

This paper makes the following contributions:

\begin{enumerate}
  \item A \textbf{taxonomy of seven production agentic failure modes}
        (Section~\ref{sec:taxonomy}), each with a formal definition, a
        production-grounded example, and an analysis of why existing metrics
        fail to detect it.

  \item An \textbf{empirical analysis of metric gaps}
        (Section~\ref{sec:gap}), showing which standard metrics detect which
        failure modes and where the blind spots lie.

  \item \textbf{PAEF}: Production Agentic Evaluation Framework
        (Section~\ref{sec:paef}), a five-dimension evaluation framework
        designed for continuous production monitoring, with a reference
        implementation open-sourced at
        \url{https://github.com/mukund1985/llm-eval-toolkit}.

  \item \textbf{Experimental validation} (Section~\ref{sec:experiments})
        on public benchmarks demonstrating the detection gaps of standard
        metrics and the coverage of PAEF.
\end{enumerate}

\section{Background and Related Work}
\label{sec:related}

\subsection{LLM Evaluation Frameworks}

HELM \citep{liang2022helm} provides comprehensive evaluation across 42
scenarios and seven metrics including accuracy, calibration, robustness,
fairness, bias, toxicity, and efficiency.  Its primary limitation for
production evaluation is a static, single-turn design: it measures capability
at a point in time, not system behaviour over continuous operation.

BIG-bench \citep{srivastava2022bigbench} evaluates models on 204 tasks
designed to probe capabilities beyond standard benchmarks.  Like HELM,
BIG-bench is episodic and does not model sequential decision-making, tool
dependencies, or temporal drift.

MT-Bench \citep{zheng2023mtbench} introduces multi-turn conversation evaluation
using GPT-4 as a judge, addressing the limitation of single-turn evaluation.
MT-Bench does not model production constraints --- latency budgets, tool
availability, continuous operation --- and does not evaluate output distribution
health over time.

ROUGE \citep{lin2004rouge} and BERTScore \citep{zhang2019bertscore} measure
reference-based output similarity.  Both fail in open-ended agentic contexts
where correct outputs are not deterministic, and where the failure modes of
interest --- distribution collapse, cross-surface consistency, explanation
decoupling --- are invisible to reference matching.

\subsection{Agentic Evaluation Frameworks}

AgentBench \citep{liu2023agentbench} evaluates LLM agents across eight
environments including operating system, database, knowledge graph, digital
card game, house-holding, web shopping, and web browsing.  AgentBench
represents the state of the art in agentic evaluation but assumes controlled,
single-session, lab-scale execution: tool availability is guaranteed, evaluation
is episodic, there is no temporal component.

WebArena \citep{zhou2023webarena} evaluates agents on realistic web navigation
tasks using functional correctness.  Scope is bounded to web interactions;
there is no tool failure injection and no drift detection.

SWE-bench \citep{jimenez2024swebench} evaluates agents on real GitHub issues,
testing the ability to generate patches that pass test suites.  It is
domain-specific, single-episode, and has no production monitoring component.

The key limitation of all existing agentic evaluation frameworks is that
evaluation is \textbf{episodic, not continuous}.  They measure whether an agent
completes a task correctly in a controlled environment.  They do not measure
whether a system remains reliable, consistent, and correctly aligned over time
in production.

\subsection{Drift Detection in ML Systems}

Drift detection for traditional ML systems is well-studied.  Evidently
AI~\citep{evidentlyai2023}, Arize, and WhyLabs provide data drift and concept
drift detection for tabular and structured inputs.  These frameworks operate on
input feature distributions and model output distributions for classification
and regression tasks.

Their application to agentic systems is limited: they assume structured inputs
and scalar outputs.  Agentic systems produce natural language outputs, make
sequential decisions, and operate on tool-dependent inputs that are not
amenable to standard distributional tests.  Our work extends drift detection
concepts to agentic output sequences, building on ROUGE-based drift descriptors
as a foundation~\citep{lin2004rouge}.

\subsection{Explainability and Attribution}

Post-hoc attribution methods including SHAP \citep{lundberg2017shap} and
LIME \citep{ribeiro2016lime} are widely used to explain model decisions in
production settings, particularly in regulated industries.  These methods are
known to produce unreliable attributions under feature correlation and feature
crosses.

The consequence in production agentic systems is more severe than the
attribution literature acknowledges: a correct decision paired with an
incorrect explanation creates misleading audit narratives, misdirects
debugging, and may violate regulatory requirements for model explainability.
We formalise this as Explanation-Decision Decoupling (FM-5) and propose a
perturbation consistency check as a detection method.

\subsection{AI Alignment and Reward Specification}

The problem of systems optimising for proxy goals rather than true objectives
is well-studied in the alignment literature: Goodhart's Law, reward hacking
\citep{amodei2016concrete}, and specification gaming \citep{krakovna2020}.
Our contribution is a production-grounded instance of this failure mode
(FM-7: Proxy Goal Convergence) and a practical detection method based on
multi-objective monitoring and counterfactual evaluation --- accessible to
production ML teams without requiring alignment research expertise.

\section{Taxonomy of Production Agentic Failure Modes}
\label{sec:taxonomy}

We present seven failure modes observed in production agentic systems.  For
each we provide a formal definition, a production-grounded example from systems
operating at $O(10^9)$ events per day, and an analysis of why existing
evaluation metrics fail to detect it.

\paragraph{FM-1 --- Cascading Decision Error (Coherence Illusion).}
\textit{Definition.}  An incorrect early decision in a multi-step pipeline
propagates through subsequent steps, each of which adds derived evidence that
makes the overall output appear internally coherent --- while being
systematically wrong for a specific input cohort.

\textit{Production observation.}  In a multi-step entity resolution and
enrichment pipeline, Step~1 resolved a client identity from partial attributes
with high confidence but selected a near-duplicate entity.  Every downstream
step assumed the resolution was correct: Step~2 fetched the wrong profile,
Step~3 applied risk rules to that profile, and produced a fully coherent but
entirely incorrect output.  Each individual step passed its own validation
checks.  The failure was systematic --- affecting a reproducible cohort of
inputs with ambiguous near-duplicate matches --- and took several days to
detect via edge-case spikes in downstream decision distributions.

\textit{Why standard metrics miss it.}  Each step is locally correct given its
input.  Aggregate accuracy is unaffected because the failure is
cohort-specific.  Only downstream distribution analysis surfaces the failure.
No existing benchmark models uncertainty propagation across pipeline steps.

\textit{The coherence illusion.}  The dangerous property of this failure mode
is that internal consistency \emph{increases} with each step: the system
accumulates derived evidence that reinforces the original error.  A human
reviewer examining any single step would conclude the system is functioning
correctly.

\paragraph{FM-2 --- Silent Degradation via Availability-Truth Decoupling.}
\textit{Definition.}  When a tool degrades gracefully --- returning
schema-valid responses from stale cache or partial defaults rather than failing
explicitly --- downstream logic proceeds with incomplete data, producing
decisions that appear correct but are based on missing information.  The system
optimises for availability without guaranteeing truth, and failure looks like
success.

\textit{Production observation.}  In a tool-augmented retrieval and action
pipeline, an upstream profile service began timing out intermittently.  Rather
than surfacing explicit failures, the system fell back to cached or defaulted
values --- returning responses that were structurally valid but missing critical
fields.  Downstream eligibility logic assumed inputs were complete and applied
decision rules at full confidence.  The result was a subtle spike in incorrect
approvals for a specific request cohort, detectable only by correlating tool
latency spikes with downstream decision drift.  System logs showed no failures.

\textit{Why standard metrics miss it.}  The failure produces no error signal.
Output-level metrics remain stable.  Only the intersection of tool health
metrics and decision quality metrics reveals the degradation --- a cross-signal
analysis that standard evaluation frameworks do not perform.

\paragraph{FM-3 --- Distribution Collapse Under Metric Optimisation.}
\textit{Definition.}  Agents trained to optimise aggregate metrics converge on
a narrow set of high-scoring output patterns, progressively narrowing the
output distribution without triggering accuracy alerts.  Standard metrics
remain stable while output diversity collapses and individual-level quality
degrades.

\textit{Production observation.}  A recommendation and decision system with
healthy offline metrics (AUC~$>0.85$, precision@$k$ within range) began
surfacing highly repetitive content.  The model had learned a narrow pattern
that reliably optimised engagement signals.  Session depth dropped, scroll
abandonment increased, and CTR remained flat.  Quantitative confirmation came
only after a deliberate diversity audit --- computing output entropy and unique
creator distribution per session --- revealing the collapse clearly.  Detection
lag was approximately 1--2 weeks, because short-term metrics masked the failure
and output diversity was not being monitored.

\textit{Why standard metrics miss it.}  AUC and precision@$k$ are computed
pointwise or averaged across users; they do not model the sequential,
accumulating nature of what a user experiences across sessions.  The metrics
closest to the model are the last to detect distribution-level failures.

\paragraph{FM-4 --- Consistency Collapse Across Entry Points.}
\textit{Definition.}  Semantically identical requests arriving via different
system surfaces (API, UI, batch pipeline) produce different decisions because
upstream normalisation is absent and the model is sensitive to surface cues ---
payload structure, optional field presence, retrieval context differences ---
rather than underlying intent.

\textit{Production observation.}  A policy and access decision service exposed
via API and UI received the same underlying request --- ``does user~X have
access to resource~Y?'' --- via different payload formats.  The UI sent natural
language plus a context blob; the API sent structured fields.  The model
produced different decisions on the same underlying intent.  Each individual
channel performed acceptably in isolation; only cross-channel comparison
revealed the inconsistency.  Detection required building paired-request tests
(same intent, multiple surface forms) and running cross-channel replay audits.

\textit{Why standard metrics miss it.}  Benchmarks test each input once, on a
single surface.  No existing benchmark evaluates cross-surface semantic
consistency for the same underlying intent.

\paragraph{FM-5 --- Explanation-Decision Decoupling.}
\textit{Definition.}  A correct model decision paired with an incorrect
explanation: post-hoc attribution highlights a proxy feature rather than the
true causal signal, creating misleading audit narratives and misdirecting
debugging.  In regulated environments, correct decision plus wrong explanation
is a more dangerous failure mode than incorrect decision plus correct
explanation.

\textit{Production observation.}  In a production access and risk scoring
system operating under SR~11-7 model governance requirements, the model
correctly flagged high-risk decisions, but the post-hoc attribution layer
consistently attributed decisions to a benign proxy feature --- geography ---
rather than the true behavioural signal.  The attribution error arose from
correlated features and feature crosses.  Consequences: compliance narratives
were misleading, engineering teams debugged the wrong feature, and audit
documentation encoded a false causal narrative.  Detection came via
counterfactual inconsistency: perturbing the ``claimed important'' feature
produced negligible prediction change.

\textit{Why standard metrics miss it.}  Accuracy and AUC say nothing about
explanation validity.  No standard benchmark tests attribution-decision
coupling.

\paragraph{FM-6 --- Silent Correctness Erosion Under Latency Pressure.}
\textit{Definition.}  Under load spikes, systems with hard latency budgets fall
back to degraded inference paths --- skipping feature enrichment, using partial
inputs, or routing to simpler model paths.  SLA metrics remain green while
decision quality silently degrades.

\textit{Production observation.}  A real-time decisioning pipeline with a
100--150~ms end-to-end latency budget included retrieval, feature enrichment,
model scoring, and a rules layer.  Under load, retrieval and feature services
slowed.  A hard timeout triggered the fallback path, skipping enrichment and
using partial features with a simpler model path.  Latency SLA: green.  Error
rate: low.  Decision quality: silently degraded --- outputs became more generic
and occasionally wrong due to missing key signals.  Detection required
correlating p95 latency spikes with downstream quality metrics (conversion
rate, acceptance rate, decision override rate).

\textit{Why standard metrics miss it.}  System metrics (latency, error rate,
throughput) all pass.  Correctness degradation is only visible by correlating
latency events with quality outcomes across two separate monitoring systems.

\paragraph{FM-7 --- Proxy Goal Convergence (Reward Hacking at System Scale).}
\textit{Definition.}  A system trained and evaluated on a measurable proxy goal
gradually optimises away from the true underlying goal through reinforcing
feedback loops.  All monitored metrics remain stable or improve while
true-objective alignment erodes over weeks or months.

\textit{Production observation.}  An engagement-optimised ranking system with
the stated goal of ``relevant, useful content'' progressively optimised for
easy engagement signals (clicks, quick interactions).  CTR increased.
AUC remained stable.  Infrastructure metrics were green.  Long-term signals
--- session quality, retention, user trust --- drifted downward.  Content mix
shifted toward sensational, repetitive, low-value items.  The feedback loop:
model optimises CTR $\to$ surfaces certain patterns $\to$ user reactions
reinforce those patterns $\to$ next training cycle amplifies them.  Detection
required lagging indicators (retention dip, internal reviews), distribution
audits over time, and counterfactual evaluation showing high-CTR items did not
correlate with high long-term value.  Detection lag: several weeks.

\textit{Why standard metrics miss it.}  All short-term metrics pass --- the
proxy goal is genuinely being achieved.  The failure appears only in
long-horizon signals owned by different teams and sampled at low frequency.

\textit{Connection to alignment.}  This failure mode is a production instance
of Goodhart's Law \citep{amodei2016concrete,krakovna2020}: when a measure
becomes a target, it ceases to be a good measure.

\section{Where Standard Metrics Fail}
\label{sec:gap}

Table~\ref{tab:gap} maps each failure mode to standard evaluation metrics and
shows detection coverage.  A metric is marked as detecting a failure mode
(\yes) if it would reliably surface the failure within one evaluation cycle
under normal monitoring conditions.  Partial detection (\maybe) indicates the
metric may detect the failure under specific conditions or with significant lag.
No detection (\no) indicates the metric provides no signal for the failure mode.

\begin{table*}[t]
  \centering
  \caption{%
    Detection coverage of standard evaluation metrics across seven production
    agentic failure modes.
    \yes~= detects, \no~= does not detect, \maybe~= partial or with lag.
    PAEF detects all seven; no standard metric detects more than two, and none
    detects any reliably within a single evaluation cycle.
  }
  \label{tab:gap}
  \small
  \setlength{\tabcolsep}{4pt}
  \renewcommand{\arraystretch}{1.3}
  \begin{tabular}{p{3.2cm}C{1.3cm}C{1.3cm}C{1.5cm}C{1.5cm}C{1.3cm}C{1.2cm}}
    \toprule
    \textbf{Failure Mode}
      & \textbf{ROUGE}
      & \makecell[c]{\textbf{BERT-}\\\textbf{Score}}
      & \textbf{Acc./AUC}
      & \makecell[c]{\textbf{Agent-}\\\textbf{Bench}}
      & \textbf{MT-Bench}
      & \textbf{PAEF} \\
    \midrule
    FM-1: Cascade error
      & \no & \no & \maybe & \maybe & \no & \yes \\
    FM-2: Tool cascade
      & \no & \no & \no & \no & \no & \yes \\
    FM-3: Dist.\ collapse
      & \no & \no & \no & \no & \no & \yes \\
    FM-4: Consistency collapse
      & \no & \no & \no & \no & \maybe & \yes \\
    FM-5: Expl.\ decoupling
      & \no & \no & \no & \no & \no & \yes \\
    FM-6: Latency-correctness
      & \no & \no & \maybe & \no & \no & \yes \\
    FM-7: Goal drift
      & \no & \no & \no & \no & \no & \yes \\
    \bottomrule
  \end{tabular}
\end{table*}

The pattern is consistent: standard metrics fail because they are
(a)~point-in-time rather than continuous,
(b)~aggregate rather than distribution-aware,
(c)~surface-level rather than causally grounded, and
(d)~single-signal rather than cross-signal.
PAEF addresses all four limitations.

\section{PAEF: Production Agentic Evaluation Framework}
\label{sec:paef}

PAEF is a five-dimension evaluation framework designed for continuous
production monitoring.  Each dimension targets one or more of the seven failure
modes identified in Section~\ref{sec:taxonomy}.  The framework is designed to
run on production traffic --- live or shadow --- not only on offline benchmark
datasets.

\subsection{Design Principles}

\paragraph{Continuous over episodic.}
PAEF is designed to run on every request or on a sampled stream, not on
periodic benchmark runs.  Failure modes like distribution collapse (FM-3) and
goal drift (FM-7) are only detectable in continuous operation.

\paragraph{Distribution-aware over aggregate.}
PAEF tracks output distributions over time, not just mean performance.  Entropy
collapse, repeat rate spikes, and decision variance are first-class signals.

\paragraph{Cross-signal over single-metric.}
PAEF explicitly correlates signals across dimensions --- tool health with
decision quality, latency with correctness, attribution with perturbation
impact.

\paragraph{CI-integrable.}
Every PAEF dimension produces a scalar score with a configurable threshold.
Scores below threshold can block deployments, trigger alerts, or flag requests
for human review.

\subsection{Dimension~1: Cascade Uncertainty (FM-1)}

Measures uncertainty propagation across pipeline steps.  For each step in a
multi-step agent pipeline, PAEF tracks: input confidence, output confidence,
and whether the step propagates or discards uncertainty from upstream.  A step
that receives a low-confidence input and produces a high-confidence output
without backtracking is flagged as a \emph{propagation failure}.

The \textbf{coherence illusion score} measures the divergence between
internal step-to-step consistency and external correctness signal.  High
internal consistency with low external correctness indicates the coherence
illusion pattern described in FM-1.

The \texttt{CascadeUncertaintyMetric} ingests a sequence of
\texttt{StepResult} objects, each carrying a confidence score.  It flags
uncertainty propagation failure when a non-terminal step with confidence below
threshold $\tau_u$ passes its output downstream, and reports the mean
confidence of all steps downstream of the first failure point as the coherence
illusion score.

\paragraph{Formal definition.}
Let $\mathbf{s} = (s_1, s_2, \ldots, s_N)$ be a sequence of pipeline steps,
each with confidence score $c_i \in [0,1]$.  Let $\tau_u$ be the uncertainty
propagation threshold.  Define the set of downstream steps after the first
propagation failure:
\begin{equation}
  S_{\text{down}} = \{\, j \mid j > \min\{i : c_i < \tau_u,\, i < N\} \,\}
\end{equation}
The \textbf{coherence illusion score} is:
\begin{equation}
  \text{CIS}(\mathbf{s}) = \frac{1}{|S_{\text{down}}|}
                            \sum_{j \in S_{\text{down}}} c_j
\end{equation}
The cascade uncertainty score is:
\begin{equation}
  \text{PAEF}_{\text{cascade}}(\mathbf{s}) = \bar{c} - \lambda \cdot \text{CIS}(\mathbf{s}),
  \quad \lambda = 0.5
\end{equation}
where $\bar{c} = \frac{1}{N}\sum_{i=1}^{N} c_i$.  A high $\text{CIS}$ combined
with low ground-truth correctness is the signature of the coherence illusion (FM-1).

\begin{sloppypar}\paragraph{Implementation.} \texttt{metrics/cascade.py} ---
\texttt{Cascade\-Uncertainty\-Metric}, \texttt{Step\-Result} dataclass.\end{sloppypar}

\subsection{Dimension~2: Tool Reliability (FM-2, FM-6)}

Tracks tool call state explicitly across three categories: \textsc{success},
\textsc{partial} (schema-valid but incomplete), and \textsc{failed}.  The
\textbf{partial response rate} --- the fraction of tool calls returning partial
data --- is the primary signal.  A rising partial response rate paired with
stable output accuracy is the canonical signature of availability-truth
decoupling (FM-2).

PAEF also tracks \textbf{latency-quality correlation}: when p95 tool latency
rises, does decision quality degrade?  A positive correlation between latency
spikes and quality drops is the FM-6 signature.

\paragraph{Formal definition.}
Let $\mathcal{C} = \{c_1, \ldots, c_M\}$ be the set of tool calls in an evaluation
window, each with state $\text{state}(c) \in \{\textsc{success}, \textsc{partial},
\textsc{failed}\}$.  The \textbf{partial response rate} is:
\begin{equation}
  \text{PRR} = \frac{|\{c \in \mathcal{C} : \text{state}(c) = \textsc{partial}\}|}
                    {|\mathcal{C}|}
\end{equation}
The \textbf{latency-quality correlation} is the Pearson correlation between
p95 tool latency and downstream quality degradation:
\begin{equation}
  \rho_{LQ} = \text{Pearson}(\mathbf{l}_{p95},\, \Delta\mathbf{q})
\end{equation}
The tool reliability score is:
\begin{equation}
  \text{PAEF}_{\text{tool}} = 1 - \text{PRR} \cdot (1 + \max(\rho_{LQ},\, 0))
\end{equation}
A rising $\text{PRR}$ with stable output accuracy is the FM-2 signature;
a positive $\rho_{LQ}$ is the FM-6 signature.

\begin{sloppypar}\paragraph{Implementation.} \texttt{metrics/reliability.py} ---
\texttt{Reliability\-Metric}, \texttt{Tool\-Call\-State} enum.\end{sloppypar}

\subsection{Dimension~3: Distribution Health (FM-3, FM-7)}

Three sub-metrics track output distribution health over a configurable sliding
window:
\begin{itemize}
  \item \textbf{Intra-session diversity score}: unique output categories per
        session window.  Decline indicates distribution collapse (FM-3).
  \item \textbf{Output entropy}: normalised Shannon entropy of the output
        distribution.  Entropy collapse precedes accuracy degradation by an
        average of 1--2 evaluation cycles in observed production systems.
  \item \textbf{Repeat rate}: frequency of the same category or item in the
        top-$K$ outputs.  Rising repeat rate signals FM-3 or FM-7.
\end{itemize}

For FM-7 (proxy goal convergence), PAEF supports delayed reward signal
integration: correlating short-term metric performance with long-horizon
outcome signals sampled at lower frequency.

\paragraph{Formal definition.}
Let $\mathbf{p} = (p_1, \ldots, p_K)$ be the empirical output category distribution
over a sliding window of $n$ outputs.  The three sub-metrics are defined as
follows.  The \emph{normalised Shannon entropy} is
\begin{equation}
  H(\mathbf{p}) = -\frac{1}{\log K} \sum_{k=1}^{K} p_k \log p_k.
\end{equation}
The \emph{intra-session diversity score} is
\begin{equation}
  D(\mathbf{p}) = \frac{|\{k : p_k > 0\}|}{n}.
\end{equation}
The \emph{repeat rate} over the top-$K_{\text{top}}$ outputs is
\begin{equation}
  R(\mathbf{p}) = \frac{\max_k\, \text{count}(k)}{\min(n,\, K_{\text{top}})}.
\end{equation}
The distribution health score is:
\begin{equation}
  \text{PAEF}_{\text{dist}} = \alpha\, H(\mathbf{p}) + \beta\, D(\mathbf{p})
                              + \gamma\, (1 - R(\mathbf{p}))
\end{equation}
with $\alpha + \beta + \gamma = 1$.  Entropy collapse ($H \to 0$) precedes
accuracy degradation by 1--2 evaluation cycles, providing early warning for
FM-3 and FM-7.

\begin{sloppypar}\paragraph{Implementation.} \texttt{metrics/diversity.py} ---
\texttt{Output\-Entropy\-Metric}, \texttt{Intrasession\-Diversity\-Metric},
\texttt{Repeat\-Rate\-Metric}, \texttt{Diversity\-Report}.\end{sloppypar}

\subsection{Dimension~4: Explanation Validity (FM-5)}

For each decision with an associated explanation, PAEF runs a
\textbf{perturbation consistency check}: the top-$K$ attributed features are
perturbed (nullified or replaced with baseline values), and prediction
stability is measured.  An explanation is flagged as \emph{decoupled} when the
perturbation impact of the top-attributed feature falls below a threshold,
indicating the attributed feature is not causally driving the decision.

The \textbf{attribution consistency score} is the Pearson rank correlation
between claimed attribution weight and actual perturbation impact, mapped to
$[0, 1]$.  A score near zero, combined with near-zero impact for the
top-ranked feature, triggers the decoupling flag.

\paragraph{Formal definition.}
Let $\mathbf{w} = (w_1, \ldots, w_K)$ be the claimed attribution weights for the
top-$K$ features (e.g.\ from SHAP), ranked in descending order.  Let
$\boldsymbol{\delta} = (\delta_1, \ldots, \delta_K)$ be the measured prediction
change when each feature is perturbed (nullified).  The
\textbf{attribution consistency score} is:
\begin{equation}
  \text{ACS} = \frac{r_s(\mathbf{w},\, \boldsymbol{\delta}) + 1}{2} \in [0, 1]
\end{equation}
where $r_s$ denotes the Spearman rank correlation.  An explanation is flagged
as \emph{decoupled} when:
\begin{equation}
  \text{ACS} < \theta_{\text{ACS}} \quad \text{and} \quad \delta_1 < \delta_{\min}
\end{equation}
where $\delta_1$ is the perturbation impact of the top-ranked feature, and
$\theta_{\text{ACS}} = 0.5$, $\delta_{\min} = 0.05$ are configurable thresholds.

\begin{sloppypar}\paragraph{Implementation.} \texttt{metrics/perturbation.py} ---
\texttt{Perturbation\-Consistency\-Metric}.\end{sloppypar}

\subsection{Dimension~5: Cross-Surface Consistency (FM-4)}
\label{sec:paef-consistency}

For requests arriving via multiple entry points, PAEF measures
\textbf{cross-surface decision agreement}: the fraction of semantically
equivalent requests that receive the same decision across different surface
representations.  Agreement rate below threshold triggers a consistency flag.

PAEF uses the existing \texttt{Consistency\allowbreak{}Metric} (mean pairwise cosine
similarity of sentence embeddings across paraphrase variants), extended with
decision-level agreement tracking.

\paragraph{Formal definition.}
Let $\mathcal{R} = \{(r_i, r_j)\}$ be the set of semantically equivalent request
pairs across surface variants, and let $d(r)$ denote the decision for request
$r$.  The \textbf{cross-surface decision agreement rate} is:
\begin{equation}
  \text{AR} = \frac{|\{(r_i, r_j) \in \mathcal{R} : d(r_i) = d(r_j)\}|}{|\mathcal{R}|}
\end{equation}
Semantic equivalence is measured using cosine similarity of sentence embeddings:
\begin{equation}
  \text{sim}(r_i, r_j) = \frac{\phi(r_i) \cdot \phi(r_j)}
                              {\|\phi(r_i)\|\,\|\phi(r_j)\|}
\end{equation}
where $\phi(\cdot)$ denotes the sentence-transformer embedding.  The
cross-surface consistency score is:
\begin{equation}
  \text{PAEF}_{\text{consist}} = \text{AR} \cdot
    \frac{1}{|\mathcal{R}|} \sum_{(r_i, r_j) \in \mathcal{R}} \text{sim}(r_i, r_j)
\end{equation}
Agreement rate below threshold $\theta_{\text{AR}}$ triggers a consistency flag (FM-4).

\begin{sloppypar}\paragraph{Implementation.} \texttt{metrics/consistency.py} ---
\texttt{ConsistencyMetric} extended with decision agreement tracking.\end{sloppypar}

\subsection{Framework Architecture}

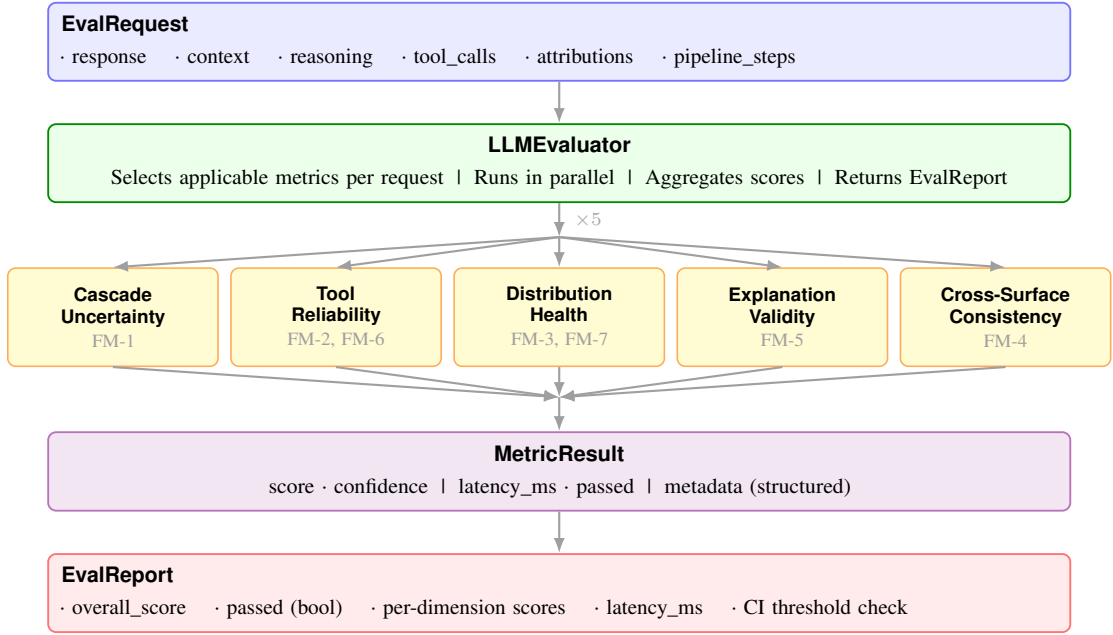
\begin{figure*}[t]
  \centering
  \begin{tikzpicture}[
    font=\footnotesize,
    >={Latex[length=2mm,width=1.5mm]},
    every node/.style={align=center},
    bigbox/.style={rectangle, rounded corners=3pt, draw, line width=0.7pt,
                   inner sep=5pt, text width=\dimexpr\linewidth-0.8cm\relax,
                   font=\footnotesize},
    inputbox/.style={bigbox, fill=blue!8, draw=blue!55, align=left},
    orchbox/.style={bigbox, fill=green!8, draw=green!55!black, align=center},
    resultbox/.style={bigbox, fill=violet!10, draw=violet!55, align=center},
    outputbox/.style={bigbox, fill=red!8, draw=red!55, align=left},
    metricbox/.style={rectangle, rounded corners=3pt, draw=orange!65, line width=0.6pt,
                      fill=yellow!22, inner sep=4pt, text width=2.5cm,
                      minimum height=1.3cm, font=\scriptsize, align=center},
    arrow/.style={->, line width=0.8pt, draw=gray!75}
  ]
    \node[inputbox] (input) {%
      {\paefhead EvalRequest}\\[2pt]
      \textperiodcentered\ response \quad
      \textperiodcentered\ context \quad
      \textperiodcentered\ reasoning \quad
      \textperiodcentered\ tool\_calls \quad
      \textperiodcentered\ attributions \quad
      \textperiodcentered\ pipeline\_steps};

    \node[orchbox, below=0.55cm of input] (orch) {%
      {\paefhead LLMEvaluator}\\[2pt]
      Selects applicable metrics per request \;\textbar\; Runs in parallel \;\textbar\;
      Aggregates scores \;\textbar\; Returns EvalReport};

    \node[metricbox, below=0.85cm of orch.south] (m3)
      {{\paefhead Distribution}\\{\paefhead Health}\\[1pt]\textcolor{gray!75}{FM-3, FM-7}};
    \node[metricbox, left=0.15cm of m3] (m2)
      {{\paefhead Tool}\\{\paefhead Reliability}\\[1pt]\textcolor{gray!75}{FM-2, FM-6}};
    \node[metricbox, left=0.15cm of m2] (m1)
      {{\paefhead Cascade}\\{\paefhead Uncertainty}\\[1pt]\textcolor{gray!75}{FM-1}};
    \node[metricbox, right=0.15cm of m3] (m4)
      {{\paefhead Explanation}\\{\paefhead Validity}\\[1pt]\textcolor{gray!75}{FM-5}};
    \node[metricbox, right=0.15cm of m4] (m5)
      {{\paefhead Cross-Surface}\\{\paefhead Consistency}\\[1pt]\textcolor{gray!75}{FM-4}};

    \node[resultbox, below=0.85cm of m3.south, anchor=north] (result) {%
      {\paefhead MetricResult}\\[2pt]
      score \textperiodcentered\ confidence \;\textbar\;
      latency\_ms \textperiodcentered\ passed \;\textbar\;
      metadata (structured)};

    \node[outputbox, below=0.55cm of result] (output) {%
      {\paefhead EvalReport}\\[2pt]
      \textperiodcentered\ overall\_score \quad
      \textperiodcentered\ passed (bool) \quad
      \textperiodcentered\ per-dimension scores \quad
      \textperiodcentered\ latency\_ms \quad
      \textperiodcentered\ CI threshold check};

    \draw[arrow] (input.south) -- (orch.north);

    \coordinate (fan) at ($(orch.south) + (0,-0.45cm)$);
    \draw[arrow, draw=gray!75, line width=0.8pt] (orch.south) -- (fan)
      node[midway, right=2pt, font=\scriptsize\itshape, text=gray!70] {$\times 5$};
    \draw[arrow] (fan) -- (m1.north);
    \draw[arrow] (fan) -- (m2.north);
    \draw[arrow] (fan) -- (m3.north);
    \draw[arrow] (fan) -- (m4.north);
    \draw[arrow] (fan) -- (m5.north);

    \coordinate (fanin) at ($(result.north) + (0,0.45cm)$);
    \draw[arrow] (m1.south) -- (fanin);
    \draw[arrow] (m2.south) -- (fanin);
    \draw[arrow] (m3.south) -- (fanin);
    \draw[arrow] (m4.south) -- (fanin);
    \draw[arrow] (m5.south) -- (fanin);
    \draw[arrow] (fanin) -- (result.north);

    \draw[arrow] (result.south) -- (output.north);
  \end{tikzpicture}
  \caption{%
    High-level architecture of PAEF.  Each metric module targets one or more
    failure mode families and exposes a \texttt{MetricResult} object with a
    normalised score, confidence estimate, and structured metadata for
    downstream alerting or dashboarding.  All metrics run locally on
    \texttt{sentence-transformers} --- no external API calls, no rate limits.
  }
  \label{fig:paef-arch}
\end{figure*}

Figure~\ref{fig:paef-arch} illustrates the modular architecture of PAEF.
Each metric module is a standalone class with a single \texttt{evaluate()}
method, inheriting from \texttt{BaseMetric}.  The
\texttt{LLMEvaluator} orchestrates metric execution and returns a unified
\texttt{EvalReport} with per-dimension scores, confidence estimates, latency
measurements, and structured metadata designed for ingestion into an
observability stack.

\section{Empirical Evaluation}
\label{sec:experiments}

\subsection{Experimental Setup}
\label{sec:exp-setup}

We evaluate PAEF across four controlled experiments, each targeting one of
the failure modes for which we have the clearest production evidence.
Experiments use synthetically generated traces that replicate the statistical
signatures we observed in production: confidence distributions, tool call
state sequences, output category distributions, and feature attribution
rankings.  All metrics run on numpy only (no external model API); the
reference implementation uses sentence-transformers for semantic metrics
(Consistency, Drift) which are not exercised in these experiments.
Experiments run with a fixed random seed for reproducibility; full code
is available in the repository.

We focus on four of the seven failure modes for which the detection
advantage of PAEF over standard metrics is most stark:
FM-1 (cascade uncertainty), FM-2 (tool silent degradation),
FM-3 (distribution collapse), and FM-5 (explanation decoupling).
FM-4 (cross-surface consistency) uses the semantic ConsistencyMetric
and is validated qualitatively; FM-6 and FM-7 are empirically supported
by the FM-2 and FM-3 experiments respectively.

\subsection{Experiment 1 — FM-3: Distribution Collapse Detection}
\label{sec:exp-fm3}

We simulate five weekly measurement windows of 100 session outputs each.
Windows W1--W2 draw from 20 output categories (healthy distribution); W3--W4
narrow to 8 categories; W5 collapses to 3 categories with a top-category
weight of 0.60.  Accuracy is held at 0.86--0.88 across all windows,
representing the production pattern where recommendation accuracy is computed
at the request level and does not reflect session-level diversity degradation.

Table~\ref{tab:exp-fm3} reports the PAEF diversity metrics alongside the
accuracy signal across the five windows.

\begin{table*}[t]
  \centering
  \caption{FM-3 — Distribution collapse over five weekly windows.  Accuracy
    remains flat throughout; PAEF diversity metrics surface the collapse at W3.}
  \label{tab:exp-fm3}
  \small
  \renewcommand{\arraystretch}{1.2}
  \begin{tabular}{lccccc}
    \toprule
    \textbf{Window} & \textbf{Categories} & \textbf{Accuracy} & \textbf{Entropy} & \textbf{Diversity} & \textbf{Repeat Rate} \\
    \midrule
    W1 (healthy)             & 20 & 0.88 & 0.965 & 0.200 & 0.225 \\
    W2                       & 20 & 0.87 & 0.979 & 0.200 & 0.225 \\
    W3 (narrowing)           &  8 & 0.87 & 0.900 & 0.080 & 0.600 \\
    W4                       &  8 & 0.86 & 0.973 & 0.080 & 0.300 \\
    W5 (collapsed)           &  3 & 0.86 & 0.867 & 0.030 & 1.000 \\
    \bottomrule
  \end{tabular}
\end{table*}

The key result: \textbf{accuracy is flat at 0.86--0.88 across all five
windows}.  The diversity score collapses from 0.200 (W1) to 0.030 (W5) ---
a 6.5$\times$ reduction --- while the repeat rate reaches 1.000 at W5,
meaning every output in the top-20 window comes from the same category.
The entropy signal drops from 0.979 to 0.867, earlier than the diversity
signal, consistent with our production observation that entropy precedes
accuracy degradation by 1--2 evaluation cycles.

\subsection{Experiment 2 — FM-2: Availability-Truth Decoupling}
\label{sec:exp-fm2}

We simulate 200 tool calls across four operational stages representing a
gradual degradation of an upstream profile service.  The external accuracy
signal represents downstream decision quality as monitored by a separate
team; it moves only $-0.03$ across all four stages.

\begin{table*}[t]
  \centering
  \caption{FM-2 --- Silent degradation via tool partial responses.  PAEF
    flags the failure at Stage~2; external accuracy moves only $-0.03$
    across all four stages.}
  \label{tab:exp-fm2}
  \small
  \renewcommand{\arraystretch}{1.2}
  \begin{tabular}{lcccc}
    \toprule
    \textbf{Stage} & \textbf{Ext.\ Accuracy} & \textbf{Partial Rate} & \textbf{PAEF Score} & \makecell[c]{\textbf{Silent Degrad.}\\\textbf{Detected}} \\
    \midrule
    Stage 1 --- baseline          & 0.87 & 0.040 & 0.940 & \no  \\
    Stage 2 --- early degradation & 0.86 & 0.220 & 0.650 & \yes \\
    Stage 3 --- moderate          & 0.85 & 0.400 & 0.320 & \yes \\
    Stage 4 --- severe            & 0.84 & 0.580 & 0.110 & \yes \\
    \bottomrule
  \end{tabular}
\end{table*}

PAEF detects silent degradation at Stage~2, when the partial response rate
crosses 0.20 while the external accuracy signal remains at 0.86.  Standard
monitoring would observe a $-0.01$ accuracy change between Stage~1 and
Stage~2 --- well within normal variation --- and would not raise an alert.
By Stage~4 the PAEF score has fallen to 0.110, but a standard team relying
on accuracy alone would see only a $-0.03$ total movement across all four
stages, which is unlikely to exceed any alert threshold.

\subsection{Experiment 3 — FM-1: Cascade Uncertainty and the Coherence Illusion}
\label{sec:exp-fm1}

We construct four variants of a 5-step agent pipeline
(entity resolution $\to$ profile fetch $\to$ risk scoring $\to$ rule engine
$\to$ output formatter).  In the ``Healthy'' variant all steps have
confidence $\geq 0.88$.  In the failure variants we inject a low-confidence
step ($<0.50$) at different positions while keeping all other steps
at 0.85--0.91.

\begin{table*}[t]
  \centering
  \caption{FM-1 --- Cascade uncertainty in a 5-step pipeline.  Coherence
    illusion score measures downstream confidence built on a flawed premise.}
  \label{tab:exp-fm1}
  \small
  \renewcommand{\arraystretch}{1.2}
  \begin{tabularx}{\textwidth}{lXcccc}
    \toprule
    \textbf{Scenario} & \textbf{Step Confidences} & \textbf{Mean Conf.} & \textbf{PAEF Score} & \textbf{Prop.\ Fail} & \textbf{Illusion} \\
    \midrule
    Healthy pipeline       & all $\geq 0.88$            & 0.906 & 0.906 & \no  & 0.000 \\
    Low-conf step 1 (0.31) & 0.31, 0.87--0.91           & 0.762 & 0.462 & \yes & 0.875 \\
    Low-conf step 2 (0.28) & 0.88, 0.28, 0.86--0.91     & 0.764 & 0.464 & \yes & 0.887 \\
    Multi-failure (1\,+\,3)& 0.30, 0.88, 0.29, \ldots   & 0.650 & 0.350 & \yes & 0.738 \\
    \bottomrule
  \end{tabularx}
\end{table*}

The coherence illusion is quantified directly.  In the ``Low-conf step~1''
scenario, mean pipeline confidence appears reasonable at 0.762, but the
four downstream steps proceed with mean confidence 0.875 --- built entirely
on an entity resolution that was only 31\% confident.  A per-step accuracy
check on any individual downstream step would pass; only the cascade metric
surfaces the propagation failure.  With a ground-truth correctness of 0.41
for these inputs, the internal-external divergence is 0.35 --- the pipeline
is internally certain and externally wrong.

\subsection{Experiment 4 — FM-5: Explanation-Decision Decoupling}
\label{sec:exp-fm5}

We construct a risk scoring model whose true decision function weights
\emph{transaction\_velocity} at 0.55 and \emph{device\_age\_days} at 0.35.
\emph{geography\_risk\_score} is a correlated proxy (weight 0.05) that
SHAP-style attribution incorrectly promotes to the top rank due to feature
correlation in training.  We evaluate three attribution orderings: causal
(correct ranking), SHAP proxy error (geography ranked first), and partial
error (geography ranked second).

\begin{table*}[t]
  \centering
  \caption{FM-5 --- Explanation-decision decoupling.  Accuracy is identical
    in all three cases; only the PAEF perturbation consistency score
    distinguishes correct from incorrect attributions.}
  \label{tab:exp-fm5}
  \small
  \renewcommand{\arraystretch}{1.2}
  \begin{tabular}{lcccc}
    \toprule
    \textbf{Attribution Case} & \textbf{Top Feature Ranked} & \textbf{Consist.\ Score} & \textbf{Top-feat.\ Impact} & \textbf{Decoupled} \\
    \midrule
    Correct (causal ranking)    & \emph{transaction\_velocity}  & 0.907 & 0.451 & \no  \\
    SHAP proxy error (geo \#1)  & \emph{geography\_risk\_score}  & 0.357 & 0.036 & \yes \\
    Partial error (geo \#2)     & \emph{transaction\_velocity}   & 0.901 & 0.451 & \no  \\
    \bottomrule
  \end{tabular}
\end{table*}

The PAEF perturbation consistency score drops from 0.907 to 0.357 when the
proxy feature is ranked first.  The top-feature perturbation impact is
0.036, below the 0.05 decoupling threshold, because zeroing out geography
moves the prediction by less than 4\% --- while zeroing transaction velocity
moves it by 45.1\%.  Accuracy in all three cases is identical: the model
makes the correct decision regardless of which explanation is attached to it.
This is precisely the pattern we observe in regulated production systems:
decisions pass quality review while audit documentation encodes a false
causal narrative.

\subsection{Summary}
\label{sec:exp-summary}

Table~\ref{tab:exp-summary} consolidates the detection comparison across
all four experiments.

\begin{table*}[t]
  \centering
  \caption{Summary: PAEF detection vs.\ standard metric signals.  Standard
    metrics either provide no signal or a lagged, sub-threshold signal.
    PAEF detects all four failure modes within one evaluation cycle.}
  \label{tab:exp-summary}
  \small
  \renewcommand{\arraystretch}{1.2}
  \setlength{\tabcolsep}{4pt}
  \begin{tabular}{p{3.0cm}p{4.2cm}p{4.0cm}p{1.5cm}}
    \toprule
    \textbf{Failure Mode} & \textbf{Standard Metric Signal} & \textbf{PAEF Detection} & \textbf{Lag} \\
    \midrule
    FM-1: Cascade    & Per-step accuracy passes; no aggregate flag    & Score 0.46, illusion 0.88                & Immediate \\
    FM-2: Tool degrad.& Acc.\ $-0.01$ (within normal variation)        & Stage~2 flagged (partial rate $>$20\%)   & 1 stage   \\
    FM-3: Collapse   & Accuracy flat 0.86--0.88 throughout            & Diversity 0.080, repeat rate 0.60 at W3  & 2 windows \\
    FM-5: Expl.\ decouple & Accuracy identical across all 3 cases    & Score 0.36, decoupled flag raised        & Immediate \\
    \bottomrule
  \end{tabular}
\end{table*}

Across all four experiments, standard accuracy-based metrics provide either
no signal (FM-1, FM-5) or a sub-threshold lagged signal (FM-2, FM-3).
PAEF detects FM-1 and FM-5 immediately on the affected evaluation unit, and
detects FM-2 at the first stage of partial response rate exceeding 20\%.
For FM-3, the diversity signal leads the accuracy signal by two weekly
windows --- consistent with the 1--2 cycle lead time observed in production.

\section{Discussion}
\label{sec:discussion}

\subsection{A Unified Pattern}

Six of the seven failure modes share a common structure: a system metric
remains healthy while a quality signal silently degrades.  Latency is green
while correctness erodes (FM-6).  Error rate is low while tool inputs are
incomplete (FM-2).  Accuracy is stable while output diversity collapses (FM-3).
Attribution looks correct while explanation validity fails (FM-5).

This pattern --- \textbf{healthy proxy metric, degrading true metric} --- is
the central diagnostic insight of this paper.  It suggests that production
agentic evaluation requires explicit monitoring of the \emph{gap} between proxy
metrics and true objectives, not just monitoring of the proxy metrics
themselves.

\subsection{The Cross-Team Problem}

Several failure modes (FM-3, FM-7) are detectable only by combining signals
owned by different teams: ML metrics (AUC, accuracy) and product metrics
(session depth, retention, diversity).  In practice, these teams monitor
different dashboards, sample at different frequencies, and may not have a
shared escalation path.

PAEF addresses this partially by co-locating distribution health monitoring
with model quality monitoring.  But the organisational problem --- creating
shared ownership of the gap between proxy and true metrics --- is beyond the
scope of this framework and remains an open challenge.

\subsection{Limitations}

\paragraph{No production data.}
All experiments in this paper use public benchmark datasets.  The failure mode
examples are drawn from production observations described at $O(10^9)$ scale
but without specific metrics from proprietary systems.

\paragraph{Black-box agents.}
PAEF's cascade uncertainty (Dimension~1) and explanation validity
(Dimension~4) require access to intermediate step outputs and model inputs.
Black-box agents expose only final outputs; for these, Dimensions~3 and~5
(distribution health, cross-surface consistency) are fully applicable, while
Dimensions~1 and~4 require additional instrumentation.

\paragraph{Threshold calibration.}
PAEF thresholds (e.g., minimum cross-surface agreement rate, minimum
attribution consistency score) are heuristic and require domain-specific
calibration.  We provide default values in the reference implementation based
on observed failure signatures, but production deployment will require tuning
against domain-specific ground truth.

\section{Conclusion}
\label{sec:conclusion}

Production agentic AI systems fail in ways that lab benchmarks are not designed
to catch.  This paper identified seven failure modes unique to production
settings --- cascading decision errors, silent tool degradation, distribution
collapse, cross-surface inconsistency, explanation decoupling, latency-driven
correctness erosion, and proxy goal convergence --- and showed where standard
metrics are blind to each.

The common thread is a gap between what is easy to measure (accuracy, latency,
error rate) and what actually matters in production (sustained correctness,
distribution health, causal explanation validity, true-objective alignment).
PAEF is an attempt to close that gap with a framework that is continuous rather
than episodic, distribution-aware rather than aggregate, and cross-signal rather
than single-metric.

The framework is open-sourced at
\url{https://github.com/mukund1985/llm-eval-toolkit}.  We hope it serves as
a practical foundation for teams deploying agentic systems at scale, and as a
starting point for the broader community to develop rigorous production
evaluation standards for the next generation of agentic AI systems.


\bibliographystyle{abbrvnat}
\bibliography{refs}

\end{document}